\documentclass[sigconf]{acmart}

\copyrightyear{2021}
\acmYear{2021}
\setcopyright{acmlicensed}
\acmConference[CIKM '21] {Proceedings of the 30th ACM International Conference on Information and Knowledge Management}{November 1--5, 2021}{Virtual Event, Australia.}
\acmBooktitle{Proceedings of the 30th ACM Int'l Conf. on Information and Knowledge Management (CIKM '21), November 1--5, 2021, Virtual Event, Australia}
\acmPrice{15.00}
\acmISBN{978-1-4503-8446-9/21/11}
\acmDOI{10.1145/3459637.3482169}
\usepackage{balance}
\usepackage{amsmath,amsfonts}
\usepackage{euler}
\usepackage{algorithmic}
\usepackage{graphicx}
\usepackage{textcomp}
\usepackage{xcolor}
\usepackage{xcolor}
\usepackage{xspace}
\usepackage[ruled, vlined, linesnumbered]{algorithm2e}
\usepackage{algorithmic}
\usepackage{graphicx}
\usepackage{amsmath}
\usepackage{amsfonts}
\usepackage{subcaption}
\usepackage{amsthm}
\usepackage{mathrsfs}
\usepackage{booktabs}
\usepackage{color}
\usepackage{hyperref}
\usepackage{lipsum}

\newcommand{\ourm}{\textsc{Reformd}\xspace}
\newcommand{\ie}{\emph{i.e.}}

\theoremstyle{definition}
\newtheorem*{problem*}{Problem Statement}%[section]

\newcommand{\cm}[1]{\mathcal{#1}}

\newcommand{\bs}[1]{\boldsymbol{#1}}

\usepackage{colortbl}

\newcommand{\nflow}{\mathcal{L}\mathcal{N}}
\newcommand{\org}{\text{org}}
\newcommand{\tgt}{\text{tgt}}

\renewcommand{\comment}[1]{}
\usepackage{xcolor}
\usepackage{multirow,amsmath,booktabs,dcolumn}
\usepackage{todonotes}
\usepackage{colortbl}
\usepackage{paralist}
\usepackage{graphicx}
\usepackage{caption}

\usepackage{mathtools}

\SetAlFnt{\footnotesize}
\usepackage{xcolor,colortbl}
\newcolumntype{a}{>{\columncolor{blue!5}}c}
\newcolumntype{b}{>{\columncolor{red!5}}c}

\begin{document}
\fancyhead{}

\title{Region Invariant Normalizing Flows for Mobility Transfer}
\author{Vinayak Gupta}
\affiliation{
  \institution{IIT Delhi}}
\email{vinayak.gupta@cse.iitd.ac.in}

\author{Srikanta Bedathur}
\affiliation{
  \institution{IIT Delhi}}
\email{srikanta@cse.iitd.ac.in}

\setlength{\abovedisplayskip}{1pt}
\setlength{\belowdisplayskip}{1pt}

\begin{abstract}
There exists a high variability in mobility data volumes across different regions, which deteriorates the performance of spatial recommender systems that rely on region-specific data. In this paper, we propose a novel transfer learning framework called \ourm, for continuous-time location prediction for regions with sparse checkin data. 
Specifically, we model user-specific checkin-sequences in a region using a marked temporal point process (MTPP) with \textit{normalizing} flows to learn the inter-checkin time and geo-distributions. Later, we transfer the model parameters of spatial and temporal flows trained on a data-rich \textit{origin} region for the next check-in and time prediction in a \textit{target} region with scarce checkin data. We capture the evolving region-specific checkin dynamics for MTPP and spatial-temporal flows by maximizing the joint likelihood of \textit{next} checkin with three channels (1) checkin-category prediction, (2) checkin-time prediction, and (3) travel distance prediction. Extensive experiments on different user mobility datasets across the U.S. and Japan show that our model significantly outperforms state-of-the-art methods for modeling continuous-time sequences. Moreover, we also show that \ourm can be easily adapted for product recommendations \ie, sequences without any spatial component.
\end{abstract}

\maketitle

\section{Introduction}\label{intro}
Recent research has shown that accurate advertisements on Points-of-Interest (POI) networks, such as Foursquare and Instagram, can achieve up to 25 times the return-on-investment~\cite{fsqstats}. Consequently, predicting the time-evolving mobility of users, \ie \emph{where} and \emph{when}, is of utmost importance to power systems relying on spatial data. 
Current approaches~\cite{locate, cho, lbsn3} overlook the temporal aspect of a recommender system as it involves modeling continuous-time checkin sequences -- which is non-trivial with standard neural architectures~\cite{rmtpp, thp, nhp}. The problem is further aggravated by the variation in volumes of mobility data across regions due to the growing awareness for personal data privacy~\cite{privacy, privacy2}. Therefore, there exists an underlying region-based data-scarcity, which further exacerbates the training of large neural models.

In recent years, Marked Temporal Point Processes(MTPP) have outperformed other neural architectures for characterizing asynchronous events localized in continuous time and are even used in a wide range of applications, including healthcare~\cite{rizoiu2}, finance~\cite{sahp, bacry}, and social networks~\cite{imtpp, nhp, thp}. Recent works that deploy MTPP for predicting user mobility patterns are either: (i) limited to predicting the time of user-location interactions rather than actual locations~\cite{chandan}, (ii) restricted to \emph{one} dataset without a foreseeable way to easily utilize external information~\cite{ank}, or (iii) disregard the opportunity to reuse trained parameters from external datasets by jointly embedding the checkin and time distributions~\cite{deepjmt}. Thus, none of these approaches can be used for designing mobility prediction models for limited data regions.

In this paper, we present \ourm(\textbf{Re}usable \textbf{F}lows f\textbf{or} \textbf{M}obility \textbf{D}ata), a novel transfer learning framework that learns spatial and temporal distribution of checkins using normalizing flows(NFs) on a checkin-rich \emph{origin} region and transfers them for efficient prediction in a checkin-scarce \emph{target} region. Specifically, we consider the series of checkins made by a user as her checkin sequence and model these sequences for all users from a region using a neural MTPP and learn the inter-checkin time interval and spatial-distance distributions as two independent NFs~\cite{flowbook, shakir}. To make the learned spatial and temporal NFs \emph{invariant} of the underlying region, we restrict our model to learn the distribution of inter-checkin time intervals and spatial distance. These features are unaffected by the network characteristics that vary across regions -- POI categories and user affinities towards these POIs. Therefore, these NFs can be easily extended for prediction in other mobility regions. The ability of NFs to provide faster sampling and closed-form training for continuous-time event sequences~\cite{intfree} make them a perfect medium to transfer mobility information. Moreover, for transferring across regions, we cluster the checkin sequences of each region, with each cluster containing checkin-sequences with \emph{similar} spatial and temporal checkin patterns and only transfer the parameters across these clusters. 

In summary, the key contributions we make in this paper via \ourm are three-fold:
\begin{inparaenum}[(i)]
\item We propose \ourm, a transfer-learning model for predicting mobility dynamics in checkin-scarce datasets by incorporating mobility parameters trained on a checkin-rich region.
\item We present a novel NF-based transfer over the MTPP that not only enables a faster sampling of time and distance features of next checkin, but also achieves high performance even with limited fine-tuning on the target region.
\item Finally, we empirically show that \ourm outperforms the state-of-the-art models by up to 20\% and 23\% for checkin-category and time prediction and can easily be extended to product recommendation datasets.
\end{inparaenum}

\vspace{-0.1cm}
\section{Related Work}\label{rw}
The key related works fall into the following categories:

\noindent \textbf{Mobility Prediction:} 
Recent sequential POI prediction models consider the checkin trajectory for each user as a sequence of events and utilize an RNN based learning~\cite{cheng2013you, cara, lbsn3} with some variants that incorporate the spatial features as well~\cite{rnnlbsn, deepmove}. Another approach \cite{tribe} is a generic model for predicting user trajectories as well as next product recommendation. Recent approaches for checkin time prediction are limited to a single dataset~\cite{deepjmt, chandan, ank}. They also model event-times as random variables rather than sequential flows and thus cannot be used for transfer across regions.

\noindent \textbf{Temporal Point Process:}
In recent years TPPs have emerged as a powerful tool to model asynchronous events localized in continuous time~\cite{daley, hawkes}, which have a wide variety of applications, e.g  information diffusion, disease modeling, finance, etc. Driven by these motivations, in recent years, there has been a surge of works on TPPs~\cite{rizoiu2, imtpp, farajtabar}. Modeling the event sequences via a neural network led to further developments including neural Hawkes process~\cite{nhp} and several other neural models of TPPs~\cite{xiao, wgantpp, fullyneural}, but cannot incorporate heterogeneous features as in a spatial networks. The approach most similar to our model is~\cite{intfree} that learns the inter-event time intervals using NFs, but ignores the spatial dynamics and is limited to a single data source. 
\vspace{-0.2cm}
\section{Problem Setup}\label{problem}
We consider the mobility records for two regions with non-overlapping locations and users, \textit{origin} and \textit{target} as $\cm{R}^{\org}$ and $\cm{R}^{\tgt}$ respectively. For any region, we represent a user trajectory as a sequence of checkins represented by $\cm{S}_k=\{e_i=(c_i,t_i, d_i) | i \in[k] , t_i<t_{i+1}, d_i<d_{i+1}\}$, where $t_i\in\mathbb{R}^+$ is the checkin time, $d_i\in\mathbb{R}^+$ is the total distance traveled, and $c_i\in \mathcal{C}$ is a discrete category of the $i$-th checkin with $\cm{C}$ as the set of all categories, and $\cm{S}_k$ denotes the first $k$ checkins. We represent the inter-checkin times and distances as, $\Delta_{t,k} = t_{k}-t_{k-1}$ and $\Delta_{d,k} = d_{k}-d_{k-1}$ respectively and model their distribution using NFs. Our goal is to capture these region invariant dynamics in origin region for mobility prediction in target region, \ie given the checkin sequence for target region, $\cm{S}^{\tgt}_K$ and the MTPP trained on origin, we aim to predict the time and category of the \textit{next} checkin, $e^{\tgt}_{K+1}$.

\vspace{-0.2cm}
\section{Model Description}\label{model}
\begin{figure}[t]
\centering
\includegraphics[width=0.9\linewidth]{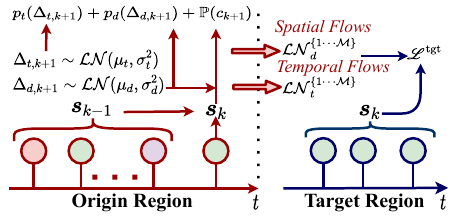}
\vspace{-0.4cm}
\caption{\label{fig:model} Architecture of \ourm with flow-based transfer between origin region (red) and target region (blue).}
\vspace{-0.2cm}
\end{figure}

We divide the working of \ourm into two parts: (i) the neural MTPP to capture mobility dynamics specific to a region, and (ii) transfer of NFs trained on the origin region to the target region. The overall schematic of \ourm is given in Figure~\ref{fig:model}.

\vspace{-0.2cm}
\subsection{Region-Specific MTPP}
We model the checkin sequences using an MTPP that we build on a recurrent neural network (RNN). The RNN is used to obtain time-conditioned vector representation of sequences as in~\cite{rmtpp, nhp, fullyneural}. Later, via these embeddings we estimate the mark distribution and inter-event time and space densities sing a three-stage architecture:

\noindent \textbf{Input stage:}
In this stage we represent the incoming checkin at index $k$, $e_k$ using a suitable vector embedding, $\bs{v}_k$ as:
\begin{equation}
\bs{v}_k = \bs{w}_{c} c_k + \bs{w}_{t} \Delta_{t,k} + \bs{w}_{d}\Delta_{d,k} + \bs{b}_v,
\end{equation}
where $\bs{w}_{\bullet}, \bs{b}_{\bullet}$ are trainable parameters and $\bs{v}_k$ denotes the vector embedding for checkin $e_k$ respectively.

\noindent \textbf{Update stage:} In this stage, we update the hidden state representation of the RNN to include the current checkin $e_k$ as:
\begin{equation}
\bs{s}_k = \tanh(\bs{G}_{s} \bs{s}_{k-1} + \bs{G}_{v} \bs{v}_{k} + \bs{g}_{t} \Delta_{t,k} + \bs{g}_{d} \Delta_{d,k} + \bs{b}_s),
\end{equation}
where $\bs{G}_{\bullet}, \bs{g}_{\bullet}, \bs{b}_{\bullet}$ are trainable parameters and $\bs{s}_k$ denotes the RNN hidden state, i.e. a \textit{cumulative} embedding for all previous checkins till the current time $t_k$.

\noindent \textbf{Output stage:}
Given the trajectory embedding $\bs{s}_k$, we predict the \textit{next} checkin time and the checkin category. Unlike~\cite{rmtpp, nhp} that learn the time distribution using the RNN hidden state, we model the \textit{density} of arrival times using a \textit{LogNormal}~\cite{intfree} flow denoted as $p_t(\Delta_{t, k+1})$ conditioned on $s_k$ as:
\begin{equation}
p_t(\Delta_{t, k+1} | \bs{s}_k) = \texttt{LogNormal} \big(\mu_t(\bs{s}_k), \sigma^2_t(\bs{s}_k)\big),
\label{eqn:flow}
\end{equation}
with $[\mu_t(\bs{s}_k), \sigma^2_t(\bs{s}_k)] = [\bs{W}_{\mu} \bs{s}_k + \bs{\mu}_t, \bs{W}_{\sigma^2} \bs{s}_k + \bs{\sigma^2}_t]$ denote the \textit{mean} and \textit{variance} of the time distribution. Such a formulation reduces model complexity, facilitates faster training and sampling in a closed-form~\cite{intfree}.

To predict the time of the next checkin, we sample the probable time difference between the current and the next checkin as $\Delta_{t, k+1} \thicksim \nflow_t\big(\mu_t(\bs{s}_k), \sigma^2_t(\bs{s}_k)\big)$, where $\nflow_t$ denotes the learned log-normal parameters. The actual time of the next checkin is the sum of the sampled time difference and the \textit{current} checkin time, $\widehat{t_{k+1}} = t_k + \Delta_{t, k+1}$. Similar to the temporal flow, we also model the inter-checkin density of spatial distances using a log-normal denoted as $p_d(\Delta_{d, k+1} | \bs{s}_k)$. We interpret this distribution as the \textit{spatial} flow for a region.

The inter-location spatial distance plays a crucial role in determining the next POI~\cite{cho, cara}. Unlike time, the distances between two checkin locations are \textit{unchanged} throughout the data. Previous approaches~\cite{rmtpp, nhp} ignore these spatial features and rely solely on the past checkin-categories. 
%This affects the prediction accuracy as the travel distance to the location is a critical element for determining POI recommendations. 
Moreover, in a sequential setting the distance that the user will travel for her next checkin is not known. Our MTPPs, being generative models, and spatial flows overcome this drawback as we can sample the probable travel distance for the next checkin from the spatial flow as $\Delta_{d, k+1} \thicksim \nflow_d\big(\mu_d(\bs{s}_k), \sigma^2_d(\bs{s}_k)\big)$. Then, for predicting the next checkin, we use the sampled distance $\Delta_{d, k+1}$ and RNN hidden state $\bs{s}_k$ via an attention weighted embedding~\cite{attention}.
\begin{equation}
\bs{s}^*_k = \bs{s}_k + \alpha \cdot \bs{w}_f \Delta_{d, k+1}, 
\label{eqn:fusion}
\end{equation}
where $\alpha, \bs{w}_f$ denote the attention weight, a trainable parameter and $\bs{s}^*_k$ denotes the \textit{updated} hidden state. We then predict the next checkin category as:
\begin{equation}
\mathbb{P}(c_{k+1} = c| \bs{s}^*_k) = \frac{\exp (\bs{V}_{s, c} \bs{s}^*_k + \bs{b}_{s, c})}{\sum_{\forall c' \in \cm{C}} \exp (\bs{V}_{s, c'} \bs{s}^*_k + \bs{b}_{s, c'})},
\end{equation}
where $\bs{V}_{s, \bullet}, \bs{b}_{s, \bullet}$ are trainable parameters and $\bullet$ denotes the entry corresponding to a category. $\mathbb{P}(c_{k+1} = c| \bs{s}^*_k)$ denotes the probability of next checkin being of category $c$ with $c \in \cm{C}$.

\noindent \textbf{Optimization:}
Given the set of all sequences $\cm{S}$ for a region $\cm{R}$, we maximize the joint likelihood for the next checkin, the log-normal density distribution of spatial, and temporal normalizing flows.
\begin{equation}
\mathscr{L} = \sum_{\forall \cm{S}}\sum_{k = 1}^{|\cm{S}|} \log \big( \mathbb{P}(c_{k+1}|\bs{s}^*_k) \cdot p_t(\Delta_{t, k+1} | \bs{s}_k) \cdot p_d(\Delta_{d, k+1} | \bs{s}_k) \big ).
\label{eqn:likelihood}
\end{equation}
where $\mathscr{L}$ denotes the joint likelihood, which we represent as the sum of the likelihoods for all user sequences. We learn the parameters of \ourm using Adam~\cite{adam} optimizer. 

\vspace{-0.3cm}
\subsection{Flow-based Transfer}
For transferring the mobility parameters across the regions, we follow the standard transfer learning procedure~\cite{transfer, inception} of training exclusively on the \textit{origin} region and then fine-tuning for the target region. However, the affinity of a user towards a POI evolves with time~\cite{cho, locate}. For example, a POI with frequent user-checkins during the summer season might not be an attractive option in winters. We include these insights by training multiple independent normalizing flows each for spatial and temporal densities. Specifically, we cluster the checkin sequences in the origin region into $\cm{M}$ equal clusters based on the \textit{median} of the occurrence times for all the checkins. Later, for each cluster of sequences, we train spatial and temporal flows independently. In this setting our net likelihood changes as to include the sum of all $\cm{M}$ likelihoods, $\mathscr{L} = \sum_{i = 1}^{\cm{M}} \mathscr{L}_i$, where $\mathscr{L}_i$ is the joint likelihood for trajectories in clusters $\cm{M}$.

\noindent As in origin region, we divide the user trajectories in the target region as well into $\cm{M}$ clusters and for trajectories in target-cluster $m^{\tgt}_i$ we attentively factor the spatial and temporal flows corresponding to origin-cluster $m^{\org}_i$. Mathematically, for the temporal flows in the target region our density of arrival times changes to:
\begin{equation}
[\mu_t(\bs{s}_k), \sigma^2_t(\bs{s}_k)]^{\tgt} = [\bs{W}_{\mu} \bs{s}_k + \bs{b}_{\mu} + \phi_t \bs{\mu}^{\org}_t, \bs{W}_{\sigma} \bs{s}_k + \bs{b}_{\sigma} + \phi_t \bs{\sigma}^{\org}_t],
\end{equation}
where $\bs{s}_k, \phi_t, \bs{\mu}^{\org}_t, \bs{\sigma}^{\org}_t$ denotes the hidden state representation for the target region, attention parameter for temporal flow and the learned flow parameters of \textit{mean} and \textit{variance} for cluster $m_i$ in origin region. Similarly, our spatial flows for target region include the origin flow parameters with an attention parameter $\phi_d$. For faster convergence, we share $\phi_t$ and $\phi_d$ across all $\cm{M}$. Other model components are same as in origin region and we maximize the joint likelihood for target region as in Equation ~\ref{eqn:likelihood}.

We highlight that our choice to divide the sequences based on \textit{median} of user trajectories rather than the individual checkin locations is driven by the following technical point: in the latter case, the \emph{net flow} --be it spatial or temporal-- would be the sum of lognormal flows for each set. Such a formulation is undesirable since the result is neither \emph{closed} nor does it remain a lognormal~\cite{logsum}, thus requiring involved techniques to approximate them~\cite{barouch1986sums}, which we would like to explore in future work. However with the current formulation, we can learn the parameters of different flows independently.

%the need to divide the flows based on \textit{median} of the whole user trajectories rather than individual checkins locations as in the latter case, the \textit{net} flow, whether spatial or temporal, would be the sum of the lognormal flows for each set. Such a formulation is neither \textit{closed} nor a lognormal~\cite{logsum} and needs inter-flow co-variance matrices. However with the current formulation, we can learn the parameters of different flows independently. 

\section{Evaluation}\label{eval}
In this section, we conduct an empirical evaluation of \ourm. Specifically, we address the following research questions.
\begin{itemize}
\item[\textbf{RQ1}] Can \ourm outperform state-of-the-art baselines for time and checkin prediction?
\item[\textbf{RQ2}] What is the advantage of transferring via normalizing flows?
\item[\textbf{RQ3}] Can we extend \ourm for non-spatial datasets? 
\end{itemize}
For evaluating mobility prediction, we consider six POI datasets from the U.S. and Japan. All our models are implemented in Tensorflow on an NVIDIA Tesla V100 GPU and are made public at \texttt{https://github.com/data-iitd/reformd}.

\begin{table}[t!]
\caption{Statistics of datasets used in our experiments. The origin region columns are followed by target regions.}
\label{tab:data}
\vspace{-0.4cm}
\centering
\resizebox{\columnwidth}{!}{
\begin{tabular}{l|cccc|cccc}
\toprule
\textbf{Property} & \textbf{NY} & \textbf{MI} & \textbf{NE} & \textbf{VI} & \textbf{TY} & \textbf{AI} & \textbf{CH} & \textbf{SA}\\
\hline
\#Users or \#Sequences ($|\cm{S}|$) & 25.6k & 6.7k & 4.1k & 6.5k & 32.1k & 10.9k & 7.5k & 11.4k\\
\#Categories ($|\cm{C}|$) & 403 & 364 & 311 & 357 & 376 & 319 & 286 & 289\\
Mean Length ($\mu_{|\cm{S}|}$) & 57.17 & 66.21 & 48.56 & 56.33 & 61.72 & 56.60 & 63.60 & 53.08\\
\bottomrule
\end{tabular}
}
\end{table}

\vspace{-0.3cm}
\subsection{Experimental Settings}
\textbf{Dataset Description:}
We use POI data from Foursquare~\cite{lbsn2vec} in United States(US) and Japan(JP) and for each country we construct 4 datasets: one with large check-in data and three with limited data. The statistics of all datasets is given in Table \ref{tab:data} with each acronym denoting the following region: (i) NY: New York(US), (ii) MI: Michigan(US), (iii) NV: Nevada(US), (iv) VI: Virginia(US), (v) TY: Tokyo(JP), (vi) CH: Chiba(JP), (vii) SA: Saitama(JP) and (viii) AI: Aichi(JP). We consider NY and TY as the origin regions and MI, NV, VI and CH, SA, AI as the corresponding target regions. For each region, we consider the time of checkin and category as event time and mark and normalize the times based on the minimum and maximum event times. We set the embedding and RNN hidden dimension to $64$ and $\cm{M} = 3$ for all our experiments. Other values for the model parameter had negligible differences.

\noindent \textbf{Evaluation Protocol:}
We split each stream of say $N$ checkins $\cm{S}_N$ into training and test set, where the training set (test set) consists of first 80\% (last 20\%) checkin. We evaluate models using standard metrics~\cite{rmtpp} of (i) mean absolute error (MAE) of predicted and actual checkin times, $\frac{1}{|\cm{S}|}\sum_{e_i\in \cm{S}}[|t_i-\widehat{t}_i|]$ and (ii) mark (checkin category) prediction accuracy (MPA), i.e, $\frac{1}{|\cm{S}|}\sum_{e_i\in \cm{S}} \#(c_i=\widehat{c}_i)$. Here $\widehat{t_i}$ and $\widehat{c_i}$ are the predicted time and category of the $i$-th checkin. Moreover, the clustering of sequences into different sets is done based solely on the training data and using these thresholds we assign clusters to sequences in the test data.

\begin{table*}[t!]
\caption{Performance of all the methods in terms of mark prediction accuracy (MPA)  and mean absolute error (MAE) across all datasets. 
%Bold fonts (underline) indicate best (second best) performer. 
Results marked \textsuperscript{$\dagger$} are statistically significant (i.e. two-sided $t$-test with $p \le 0.1$) over the best baseline.}
\vspace{-0.4cm}
\centering
\resizebox{\textwidth}{!}{
    \begin{tabular}{l|cccccc|cccccc}
\toprule
& \multicolumn{6}{c|}{\textbf{Mark Prediction Accuracy (MPA)}} & \multicolumn{6}{c}{\textbf{Mean Absolute Error (MAE)}} \\ \hline 
$\cm{R}^{\org} \rightarrow \cm{R}^{\tgt}$ & NY $\rightarrow$ MI & NY $\rightarrow$ NE & NY $\rightarrow$ VI & TY $\rightarrow$ AI & TY $\rightarrow$ CH & TY $\rightarrow$ SA & NY $\rightarrow$ MI & NY $\rightarrow$ NE & NY $\rightarrow$ VI & TY $\rightarrow$ AI & TY $\rightarrow$ CH & TY $\rightarrow$ SA\\ \hline
NHP~\cite{nhp} & 0.1745 & 0.1672 & 0.1348 & 0.2162 & 0.4073 & 0.3820 & 0.0920 & 0.1710 & 0.1482 & 0.1146 & 0.1217 & 0.1288 \\
RMTPP~\cite{rmtpp} & 0.1761 & \underline{0.1684} & 0.1377 & 0.2293 & \underline{0.4250} & 0.4036 & \underline{0.0817} & \underline{0.1581} & \underline{0.1360} & \underline{0.1058} & \underline{0.1162} & \underline{0.1205} \\
SAHP~\cite{sahp} & 0.1587 & 0.1529 & 0.1303 & 0.1968 & 0.3864 & 0.3943 & 0.1132 & 0.1958 & 0.1705 & 0.1671 & 0.1696 & 0.1574\\
THP~\cite{thp} & \underline{0.1793} & 0.1545 & \underline{0.1493} & \underline{0.2361} & 0.4229 & \underline{0.4057} & 0.0983 & 0.1735 & 0.1652 & 0.1445 & 0.1426 & 0.1468\\
\ourm & \textbf{0.2159}\textsuperscript{$\dagger$} & \textbf{0.1868}\textsuperscript{$\dagger$} & \textbf{0.1631}\textsuperscript{$\dagger$} & \textbf{0.2588}\textsuperscript{$\dagger$} & \textbf{0.4474} & \textbf{0.4208} & \textbf{0.0672}\textsuperscript{$\dagger$} & \textbf{0.1317}\textsuperscript{$\dagger$} & \textbf{0.1089}\textsuperscript{$\dagger$} & \textbf{0.0858}\textsuperscript{$\dagger$} & \textbf{0.0897}\textsuperscript{$\dagger$} & \textbf{0.0973}\textsuperscript{$\dagger$}\\\hline
$\Delta$ (\%) & 20.41 & 10.92 & 9.24 & 9.61 & 5.27 & 3.72 & 17.74 & 16.69 & 19.92 & 18.90 & 22.80 & 19.25\\
\bottomrule
\end{tabular}
}
\vspace{-0.4cm}
\label{tab:main}
\end{table*}

\vspace{-0.3cm}
\subsection{Baselines}
We compare the prediction performance of \ourm with the following state-of-the-art methods: 
\begin{asparaitem}[]
\item \textbf{RMTPP~\cite{rmtpp}}: A recurrent neural network that models time-differences to learn a representation of the past events.
\item \textbf{NHP~\cite{nhp}}: Models an MTPP using continuous-time LSTMs for capturing the temporal evolution of sequences.
\item \textbf{SAHP~\cite{sahp}}: A self-attention model to learn the temporal dynamics using an aggregation of historical events. 
\item \textbf{THP~\cite{thp}}: Extends the transformer model~\cite{transformer} to include the \textit{conditional} intensity of event arrival and the inter-mark influences.
\end{asparaitem}
We omit comparison with other continuous-time models~\cite{fullyneural, intfree, wgantpp, xiao, hawkes} as they have already been outperformed by these approaches. 

\begin{table}[b]
\caption{Prediction performance of all the methods for product recommendation in Amazon datasets. Results marked \textsuperscript{$\dagger$} are statistically significant as in Table~\ref{tab:main}.}
\vspace{-0.4cm}
\centering
\resizebox{\columnwidth}{!}{
    \begin{tabular}{l|cc|cc}
\toprule
& \multicolumn{2}{c|}{\textbf{Mark Prediction Accuracy}} & \multicolumn{2}{c}{\textbf{Mean Absolute Error}} \\ \hline 
$\cm{R}^{\org} \rightarrow \cm{R}^{\tgt}$ & DM $\rightarrow$ AP & DM $\rightarrow$ BY & DM $\rightarrow$ AP & DM $\rightarrow$ BY\\ \hline
NHP~\cite{nhp}& 0.8773 & 0.5711 & 0.0903 & 0.1795\\
RMTPP~\cite{rmtpp} & 0.8975 & 0.5530 & \underline{0.0884} & \underline{0.1758}\\
SAHP~\cite{sahp} & 0.8931 & 0.5517 & 0.1439 & 0.2214\\
THP~\cite{thp} & \underline{0.9084} & \underline{0.5879} & 0.1253 & 0.2035\\
\ourm & \textbf{0.9129} & \textbf{0.6035} & \textbf{0.0756}\textsuperscript{$\dagger$} & \textbf{0.1564}\textsuperscript{$\dagger$} \\ \hline
$\Delta$ (\%) & 0.49 & 2.65 & 14.47 & 11.03\\
\bottomrule
\end{tabular}
}
\label{tab:item}
\end{table}

\begin{figure}[t]
\centering
\begin{subfigure}[b]{0.45\columnwidth}
\includegraphics[width=\linewidth]{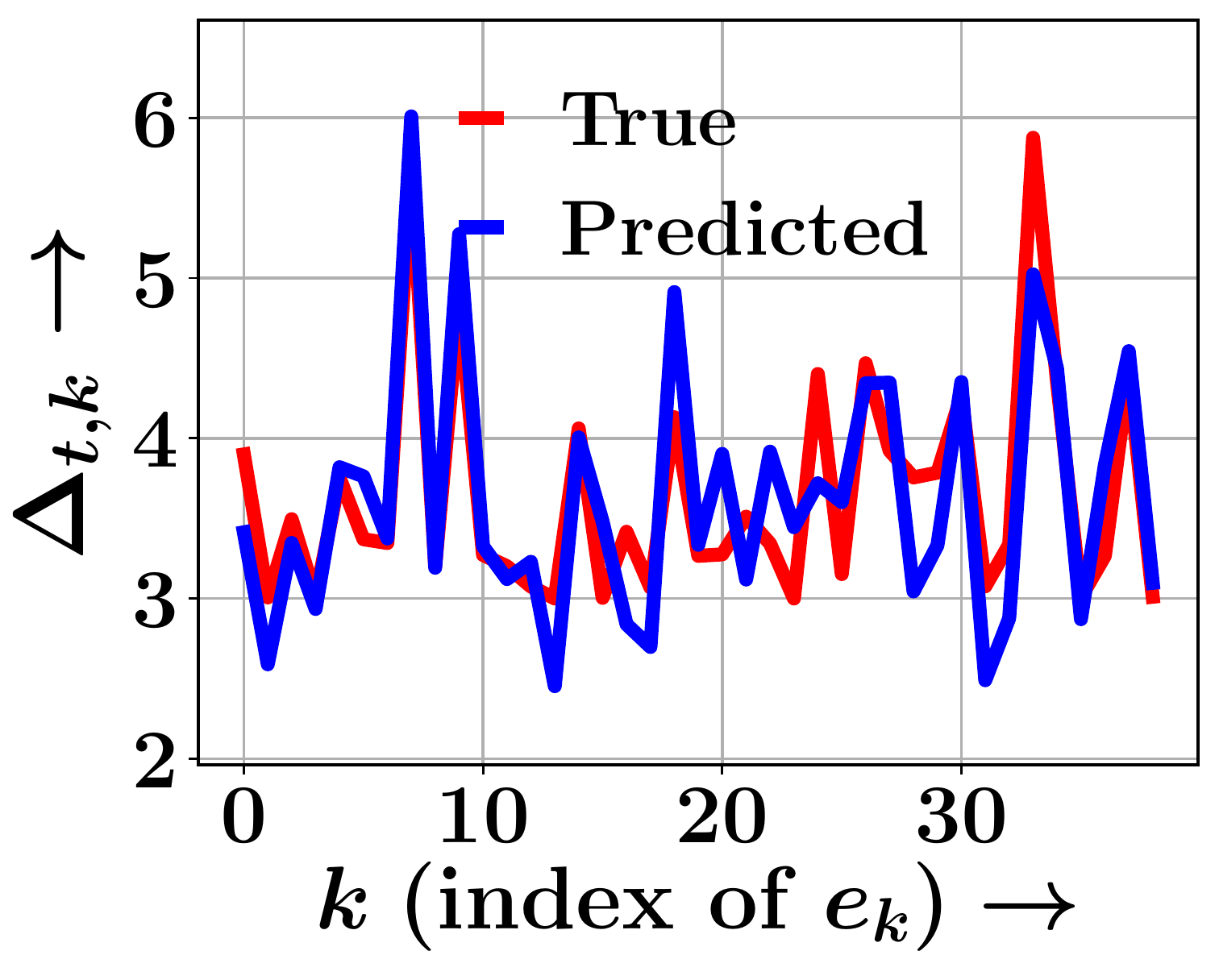}
\vspace{-0.6cm}
\caption{Virginia}
\end{subfigure}
\begin{subfigure}[b]{0.45\columnwidth}
\includegraphics[width=\linewidth]{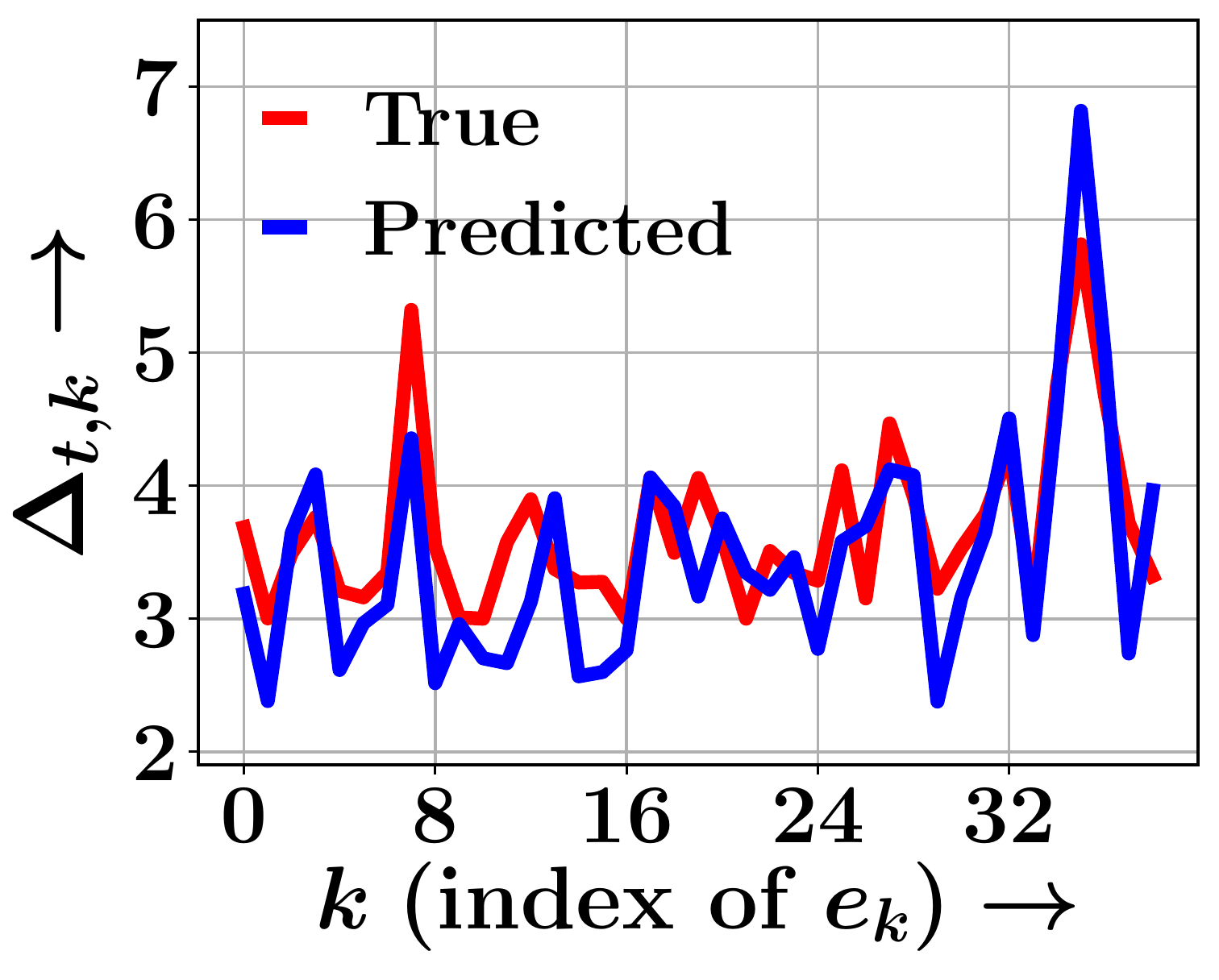}
\vspace{-0.6cm}
\caption{Aichi}
\end{subfigure}
\vspace{-4mm}
\caption{\label{fig:qualitative} Real life \textit{true} and \textit{predicted} inter-arrival times $\Delta_{t,k}$ of different events $e_k$ for (a) Virginia and (b) Aichi.}
\end{figure}

\subsection{Prediction Performance (RQ1)}
We report the prediction performance of different methods across our target datasets in Table~\ref{tab:main} and make the following observations:
\begin{asparaitem}[$\bullet$]
\item \ourm consistently yields the best performance on all the datasets. In particular, it improves over the strongest baselines by 10\% and 19\% for category and time prediction respectively. These results indicate the importance of spatial and temporal flow-based transfer from external data for prediction in limited-data regions.
\item RMTPP~\cite{rmtpp} is the second-best performer in terms of MAE of time prediction almost for all the datasets. For some datasets THP~\cite{thp} outperforms RMTPP for mark category prediction. However \ourm significantly outperforms it across all metrics. 
\end{asparaitem}
%Recent research~\cite{li_rl} has shown that NHP~\cite{nhp} and RMTPP show similar performances subjected to extensive to parameter tuning for NHP, which however is beyond scope of this paper.

\textbf{Qualitative Analysis:} We also perform a qualitative analysis to demonstrate how \ourm is able to model the checkin time distribution. For this, we plot the actual inter-checkin time differences and the difference time predicted by \ourm in Figure~\ref{fig:qualitative} for Virginia and Aichi datasets. From the results we note that the predicted inter-arrival times closely match with the true inter-arrival times and \ourm is even able to capture large time differences (peaks). For brevity, we omit the results for other datasets.

\textbf{Runtime:} For all datasets, the times for training on the origin and later on target regions is within 3 hours, thus are within the range for practical deployment.

\vspace{-0.2cm}
\subsection{Transfer Advantage (RQ2)}
\begin{figure}[t]
\centering
\begin{subfigure}[b]{0.45\columnwidth}
\includegraphics[width=\linewidth]{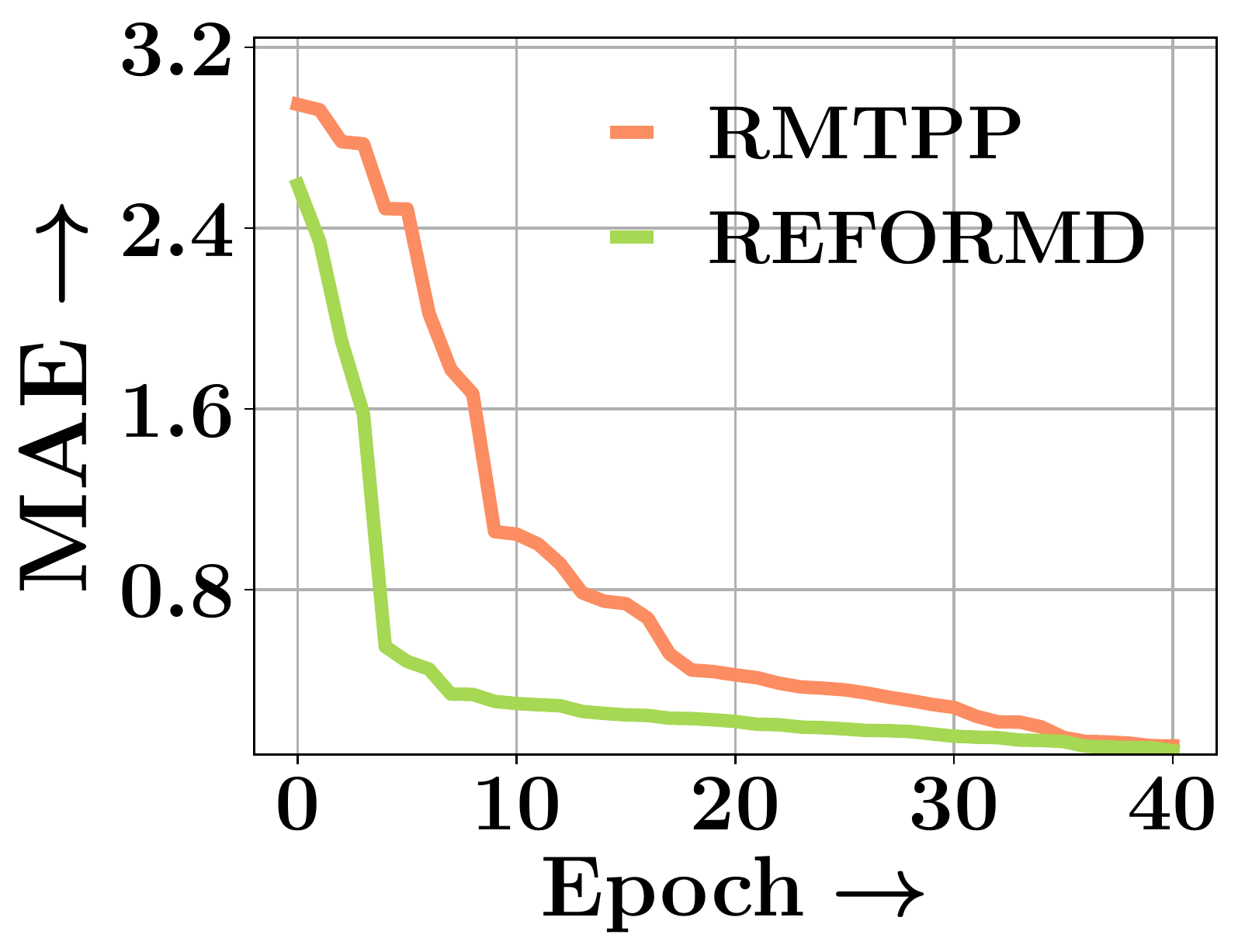}
\vspace{-0.6cm}
\caption{Virginia}
\end{subfigure}
\begin{subfigure}[b]{0.45\columnwidth}
\includegraphics[width=\linewidth]{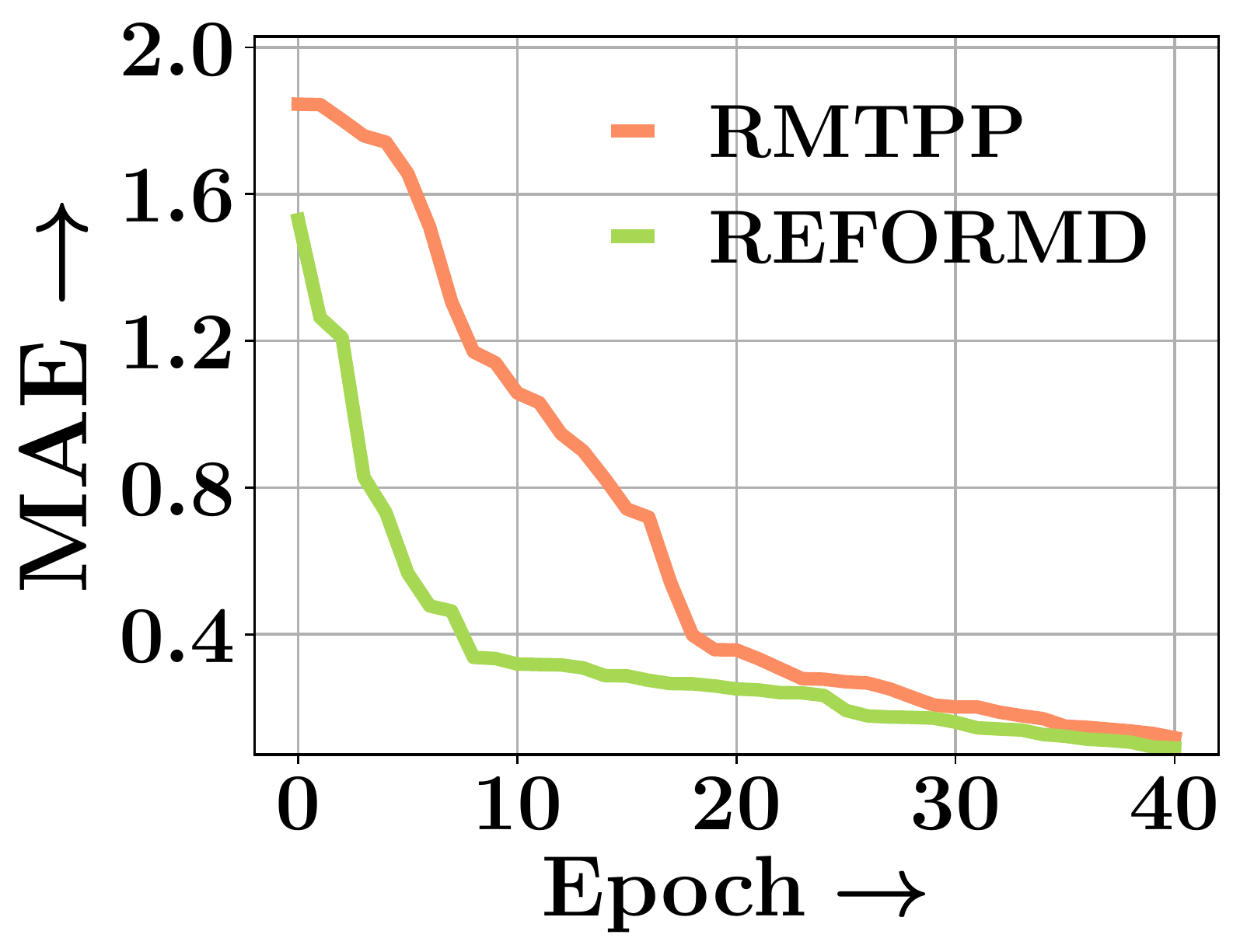}
\vspace{-0.6cm}
\caption{Aichi}
\end{subfigure}
\vspace{-4mm}
\caption{\label{fig:epochs} Training curves of \ourm and RMTPP for time prediction with \textit{best} MAE for (a) Virginia and (b) Aichi.}
\end{figure}

\ourm outperforms other baselines and also brings exhibits a key feature of \textit{transfer} learning, i.e. quick parameter learning~\cite{inception, transfer}. We highlight this by plotting the time prediction error (MAE) corresponding to the epochs trained on the target region for \ourm and the best time prediction model, i.e. RMTPP. Figure~\ref{fig:epochs} summarizes the results for Virginia and Aichi. We note that \ourm exhibits faster convergence than RMTPP for both datasets. More specifically, the flow-based transfer procedure of \ourm can outperform most baselines even with a fine-tuning of a few epochs. The results also highlight the stable learning procedure of \ourm. 

\vspace{-0.2cm}
\subsection{Product Recommendation (RQ3)}
We further evaluate the performance of \ourm in product recommendation, i.e. without spatial coordinates. Consequently, we use purchase records for three item categories from Amazon~\cite{julian}, namely Digital Music(DM), Appliances(AP) and Beauty(BY). For each item we use the user reviews as the events in a sequence with the time of the written review as the event time and the rating (1 to 5) as the corresponding mark. As in this case, we do not have a spatial density function $p_d(\Delta_{d, k+1})$, we change the \textit{fusion} equation \ref{eqn:fusion} to include the predicted time of next purchase as:
\begin{equation}
\bs{s}^*_k = \bs{s}_k + \alpha \cdot \bs{w}_f \Delta_{t, k+1}, 
\end{equation}
We consider Digital Music($|\cm{S}| = 12k$) as origin and Appliances($|\cm{S}| = 7k$) and Beauty($|\cm{S}| = 6k$) as target. From the results in Table ~\ref{tab:item}, we note that even in the absence of spatial flows, \ourm outperforms other baselines across all metrics.
%Interestingly, for product recommendations as well we see a similar trend as in spatial datasets with RMTPP outperforms other baseline models in terms of time prediction and THP~\cite{thp} and RMTPP perform competitively for mark prediction.

\vspace{-0.2cm}
\section{Conclusion and Future Work}
In this paper, we present \ourm, a novel method for transferring mobility knowledge across regions by sharing the spatial and temporal NFs for continuous-time checkin prediction. As a future work, we plan to incorporate meta-learning based transfer~\cite{maml}.

\clearpage
\bibliographystyle{ACM-Reference-Format}
\balance
\bibliography{main}
\end{document}